\documentclass{article}

\usepackage[preprint]{corl_2023} %
\usepackage{amsmath}
\usepackage{amssymb}
\usepackage{tabularx}
\usepackage{mathtools}
\usepackage{amsthm}
\usepackage{enumitem}
\usepackage{booktabs}
\usepackage[usernames,dvipsnames,svgnames,table]{xcolor}
\usepackage[export]{adjustbox}
\usepackage{wrapfig}
\usepackage{enumerate}
\usepackage{pifont}
\usepackage{fontawesome} %
\usepackage{graphicx}

\usepackage[capitalize,noabbrev]{cleveref}
\theoremstyle{plain}

\theoremstyle{definition}

\theoremstyle{remark}

\newcommand{\algoName}{\text{MT-ACT}}
\newcommand{\agentName}{\text{\textit{RoboAgent}}}
\newcommand{\roboset}{\texttt{RoboSet}}

\newcommand{\ntraj}{7,500~}
\newcommand{\ntask}{38~}
\newcommand{\nskill}{12~}
\newcommand{\nactivity}{6~}

\newcommand{\extweblink}{{\footnotesize{\faExternalLink}}}

\definecolor{ms_red}{RGB}{167, 22, 34}
\definecolor{activity-color}{RGB}{188, 75, 81}

\def\shownotes{1}  %
\ifnum\shownotes=1
\newcommand{\authnote}[2]{{$\ll$\textsf{\footnotesize #1 notes: #2}$\gg$}}
\else
\newcommand{\authnote}[2]{}
\fi

\usepackage[textsize=tiny]{todonotes}

\title{RoboAgent: \\ Generalization and Efficiency in Robot Manipulation via Semantic Augmentations and Action Chunking}

\author{
  Homanga Bharadhwaj$^{*,1,2}$ \And  Jay Vakil$^{*,2}$\And Mohit Sharma$^{*,1}$\AND Abhinav Gupta$^{1}$ \And  Shubham Tulsiani$^{1}$ \And  Vikash Kumar$^{1,2}$ \AND
  $^1$ Carnegie  Mellon University, 
  $^2$ FAIR-MetaAI  \\
  \\
  \texttt{\url{https://robopen.github.io/}\extweblink}
}
\begin{document}
\maketitle

\begin{figure}[h!]
\vspace{-1em}
    \centering
\includegraphics[trim={0cm 2cm 0 1cm},clip, width=0.95\textwidth]{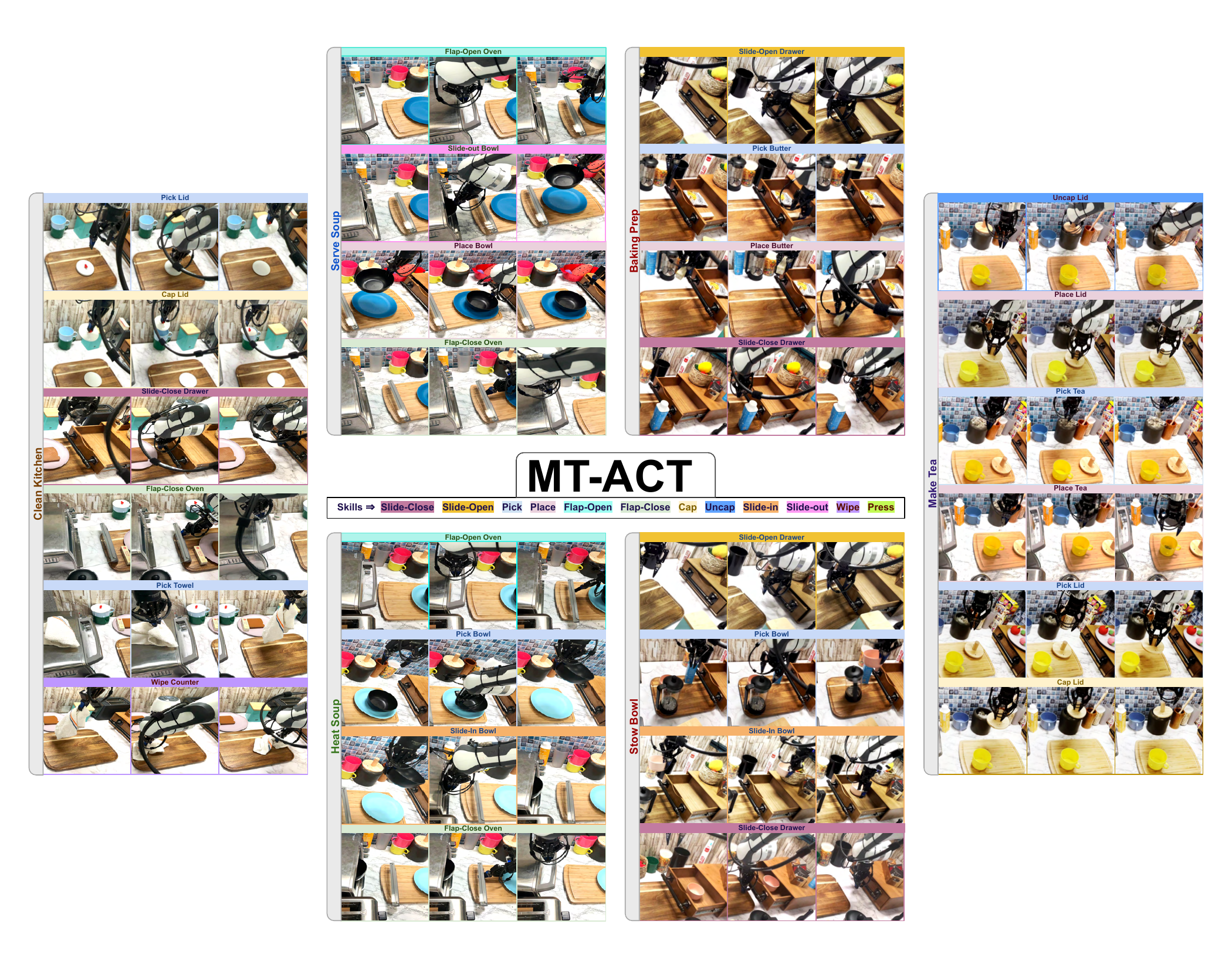}
    \caption{\footnotesize{A glimpse of the diverse manipulation capabilities of \agentName -- a single agent capable of \nskill manipulation skills across \ntask tasks encompassing \nactivity activities. For videos, visit:  \footnotesize{\url{robopen.github.io/}\extweblink}}}
    \label{fig:aug}
\end{figure}
\vspace{-1em}
\begin{abstract}
The grand aim of having a single robot that can manipulate arbitrary objects in diverse settings is at odds with the paucity of robotics datasets. Acquiring and growing such datasets is strenuous due to manual efforts, operational costs, and safety challenges. A path toward such an universal agent would require a structured framework capable of wide generalization but trained within a reasonable data budget.  
In this paper, we develop an efficient system (\agentName) for training universal agents capable of multi-task manipulation skills using (a) \textit{semantic augmentations} that can rapidly multiply existing datasets and (b) \textit{action representations} that can extract performant policies with small yet diverse multi-modal datasets without overfitting. In addition, reliable task conditioning and an expressive policy architecture enable our agent to exhibit a diverse repertoire of skills in novel situations specified using language commands. Using merely 7500 demonstrations, we are able to train a single agent capable of \nskill unique skills, and demonstrate its generalization over \ntask tasks spread across common daily activities in diverse kitchen scenes. On average, \agentName~ outperforms prior methods by over 40\% in unseen situations while being more sample efficient and being amenable to capability improvements and extensions through fine-tuning.
\end{abstract}

\section{Introduction}
\vspace*{-0.5em}

Training a robot manipulator with multiple skills requires exposure to diverse experiences and the ability to acquire skills from a diverse data corpus.
To collect such a multi-skill data corpus in the real world requires substantial effort and suffers from high operational costs and safety challenges. Given the expense, efficiency in robot learning paradigms is necessary for real-world training and deployment. While there are recent efforts in scaling real-world robotic datasets despite these challenges \cite{ebertbridge,dasari2020robonet,mandlekar2018roboturk}, efficiency seems to be overlooked in the attempts to scale~\cite{mtopt,rt1,bcz,kalashnikov2018qt}.

With the acknowledgment that robot learning will generally benefit as the scale of the robotics dataset grows, the focus of our work is on investigating generalization in developing capable agents under a \emph{given data budget}. We restrict ourselves to a dataset with \ntraj robot manipulation trajectories (an order of magnitude less than related works~\cite{rt1}) containing a diverse collection of manipulation skills across different tasks. As a robot under deployment in real environments like homes, hospitals, etc., will always find itself in unseen scenarios, we set out to develop the most capable agent with an emphasis on its \textit{ability to generalize to novel situations within this data budget}.

At first sight, wide generalization with a data budget seems like wishful thinking - while it's possible to provide large representation capabilities to the agent's policy, scaling without data diversity will likely lead to overfitting and no generalization. Our insight is twofold: (1) We ensure sufficient coverage of different skills in different scenarios in a dataset (of 7500 trajectories) we collect through teleoperation. The collected dataset is diversified offline, at no extra human/robot cost, via semantic augmentations ~\cite{genaug, cacti, rosie} to aid generalization in novel situations. (2) We train a language-conditioned manipulation policy with \textbf{\algoName} -- multi-task action-chunking transformers capable of handling the multi-modal data distribution. The architecture leverages the fact that robot movements are temporally correlated, by predicting action chunks~\cite{act} instead of per-step actions, leading to smoother behaviors and mitigation of covariate shift commonly observed in the low data imitation learning regime.

Overall, we emphasize that the data efficiency lessons we present are \emph{general} and will help in achieving generalizable agents independent of the available data budget. Building on these insights, we make the following contributions: \begin{itemize}[leftmargin=*]
\itemsep0em
 \item We present an efficient method \algoName{} designed to recover \textbf{generalist agents on a data budget}. \algoName{} leverages data multiplication via semantic augmentations and action representations to drive efficiency gains in low-data settings. 

 \item \algoName's architecture can effectively ingest multi-modal trajectory data to recover \agentName~ -- a single policy that can perform a diverse set of tasks through language instructions. Through extensive real-world experiments, we show \agentName~ is \textbf{capable of exhibiting \nskill manipulation skills}.
 
 \item We perform extensive generalization studies to demonstrate that \algoName{} is 40 $\%$ more performant than alternatives, exhibits much \textbf{superior generalization to diverse novel scenarios}, and is amenable to \textbf{improvements and extensions during deployment through fine-tuning}.
 
 \item We meticulously recorded all the data collected during the course of the project which we are open-sourcing as part of \roboset~ - one of the \textbf{largest open-source robotics dataset} on commodity hardware. It contains high-quality human teleOp trajectories spanning a balanced distribution of 12 skills across 38 tasks in diverse kitchen scenes.

\end{itemize}

\vspace*{-0.5em}
\section{Related Work}
\vspace*{-0.5em}

\begin{figure}[t]
    \centering
    \includegraphics[trim={1cm 1.8cm 1cm 1.8cm},clip, width=0.99\textwidth]{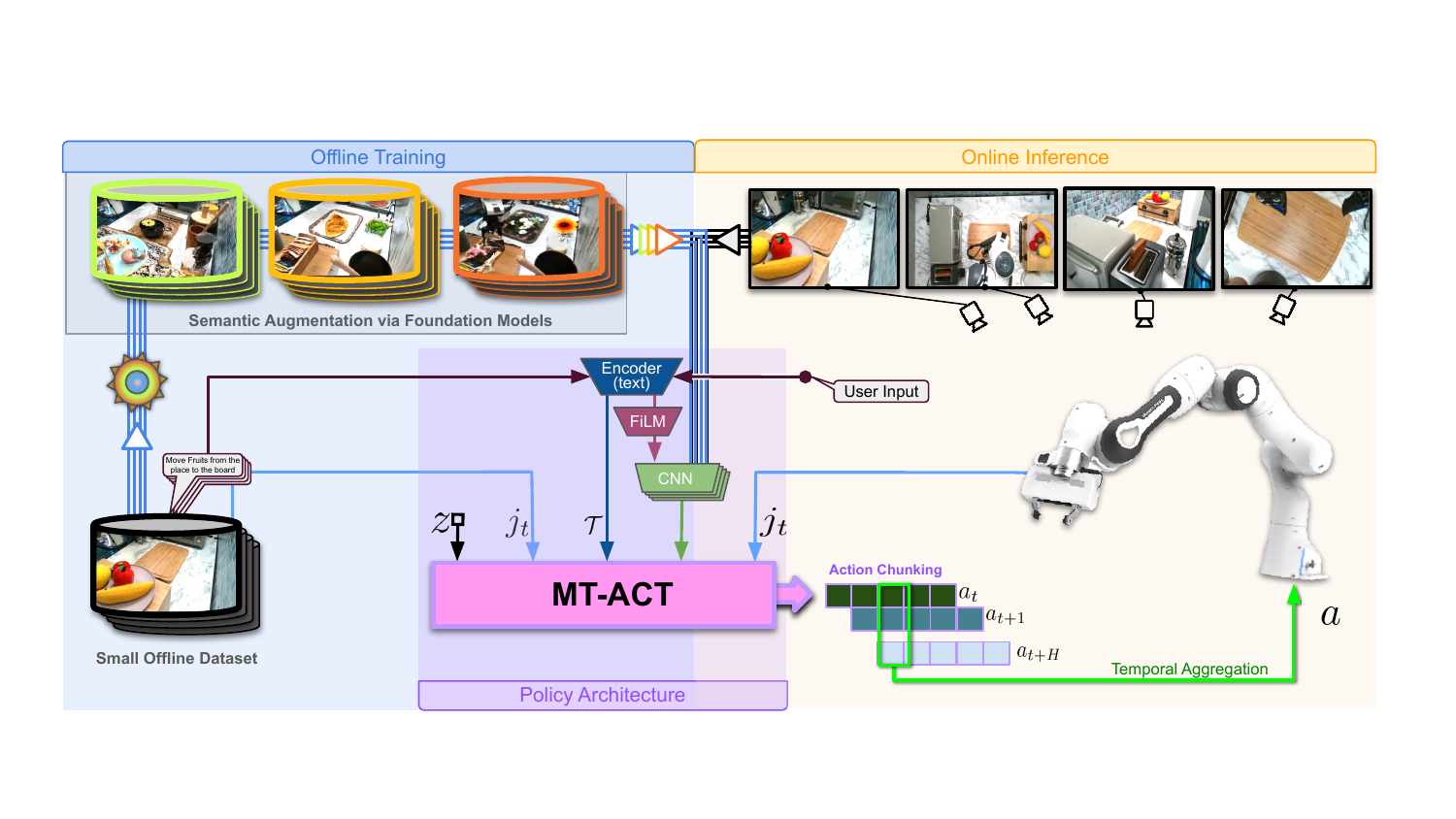}
    \caption{\footnotesize{Two stage framework: [Left] \textbf{Semantic augmentation} stage diversifies the robot data offline using inpainting augmentations at no extra human/robot cost. [Right] \textbf{Policy learning} stage trains language-conditioned policy using \algoName{} -- multi-task action-chunking transformers -- which leverages efficient action representations for ingesting multi-modal multi-task data into a single multi-skill multi-task policy.}}
    \label{fig:mainfig}
    \vspace*{-1em}
\end{figure}

\textbf{Frameworks for Scaling Robot Learning.} Given the cost of supervision in robot learning, self-supervised learning \cite{agpinto, lynch2020learning, berscheid2019improving} methods leveraging large unlabeled datasets have been a dominant paradigm in efforts towards building general-purpose agents.
Large-scale simulations \cite{metaworld, james2020rlbench, mittal2023orbit, zhu2020robosuite} have also been leveraged with the hope of learning a general multi-task policy for diverse tasks \cite{gato, vima, schrittwieser2020mastering, espeholt2018impala, sodhani2021multi, kaiser2019simple} first and then transferring it to the real world via sim2real\cite{tobin2017domain,shridhar2022cliport,handa2023dextreme,robocat}. 
However, many existing multi-task RL works focus on narrow domains in simulation~\cite{espeholt2018impala, song2019v}, and those in the real-world show limited generalization and task diversity~\cite{mtrf,cacti}.
While other works \cite{gato, vima,gradientsurgery} focus on multi-task settings in diverse scenarios, they restrict to evaluating trained policies mostly in simulation. By contrast, our work focuses on a large, diverse set of real-world manipulation tasks. Recently, many works have used imitation learning with large-scale real-world robot tele-operation datasets of high quality~\cite{dasari2020robonet, mandlekar2018roboturk, mandlekar2022matters, ebertbridge, jang2022bc,dasari2022rb2} .
While early works collect limited real-world data \cite{mandlekar2018roboturk, jang2022bc},
more recent approaches \cite{ebertbridge, rt1,kalashnikov2018qt} collect much larger datasets. In fact, \cite{rt1} gathers, possibly, the largest dataset ($\approx 130K$ demonstrations) outside bin and place settings and shows impressive generalization with skills learned using this data.
Our work is similar in spirit, \emph{i.e.}, we focus on real-world manipulation tasks and aim to learn a multi-task policy using \emph{limited} real-world demonstrations.
However, unlike \cite{ebertbridge}, we avoid toy environment setups and focus on realistic real-world kitchen setups with clutter and multiple feasible tasks in a scene.
Additionally, our agents exhibit a much greater diversity of skills than \cite{rt1,kalashnikov2018qt,robocat} while being trained only on 7.5k trajectories, as opposed to 135k in \cite{rt1}. Importantly, we collect our data with commodity hardware (see~\autoref{fig:setup}) and are making it readily available to robotic researchers worldwide.

 \textbf{Alternate Data Sources in Robotics.}
Recent successes of large-scale self-supervised approaches within both language and vision communities have showcased the advantage of large-scale data. Many recent works propose using pre-trained visual representations trained primarily on non-robot datasets~\cite{ego4d,imagenet}, for learning control policies \cite{r3m, moco, shridhar2022cliport,majumdar2023we, shah2021rrl}.  Most of these works focus on single-task settings \cite{r3m, moco, sharmalossless, hansen2022pre},\ or in simulated robot environments~\cite{shridhar2022cliport,majumdar2023we}. 
Given the inherently large cost of collecting real-world robotics datasets, many works have focused on using alternate data sources such as language~\cite{tellex2011understanding, lynch2020language, stepputtis2020language, saycan}, human videos \cite{nguyen2018translating, bharadhwajvisual, zhou2021manipulator,shao2021concept2robot,shaw2023videodex,bharadhwaj2023zero,bahl2022human,bahl2023affordances}, and generative augmentations \cite{rao2020rl, kapelyukh2022dall, cacti, genaug, rosie}.
Our work is most similar to the latter set of works, some of which use diffusion models to generate augmentations for data collected in the real world.
However, unlike some prior works \cite{cacti, genaug} our approach is fully automatic. We do not need segmentation masks \cite{cacti} or object meshes \cite{genaug} for generating augmentation data. Overall, our work is most similar to \cite{rosie} which adapts a pre-trained open-world object detection model \cite{minderer2022simple} for generating segmentations that are used with text-guided diffusion models to generate augmentations. However, our approach does not require any further fine-tuning of a separate module for open-vocabulary segmentation and language grounding. More importantly, we further investigate scaling laws with respect to semantic data augmentations to demonstrate the favorable impact of augmentations in aiding test-time generalization to unseen scenarios.

\section{\algoName: Multi-Task Action Chunking Transformer}

To learn generalizable manipulation policies, it is essential for robots to be exposed to rich and diverse experiences, encompassing a wide range of skills and contextual variations. 
However, operational costs and real-world challenges in collecting such extensive datasets pose a practical limit on their overall size. Our goal is to address these limitations by developing \textit{a paradigm that can learn effective multi-task agents under a limited data budget}. 
Our approach consists of two stages (\autoref{fig:mainfig}):

\textbf{Semantic Augmentation} -- the first stage multiplies the pre-collected dataset by creating a diverse collection of semantic augmentations over the existing robot's experiences. These semantic augmentations recreate a particular robot demonstration into several demonstrations, each with a different semantic context (objects, textures, backgrounds, etc), at no extra robot or human cost. Such data diversification incorporates real-world semantic priors into the multi-task manipulation agents preparing them to account for out-of-distribution scenarios they might encounter during deployment.

\textbf{Policy Learning} --  the second stage learns robust skills from limited skill data, adapting design choices from previously limited single-task settings to the context of large-scale generalization in multi-task multi-scene manipulation tasks with diverse skills.
We develop \algoName{} -- a language-conditioned novel policy architecture to train robust agents capable of recovering multiple skills from multi-modal datasets. To model the diverse multi-modal multi-task augmented datasets, we employ a Conditional Variational Autoencoder (CVAE) \cite{kingma2013vae} to identify action distribution modes. This enables us to fit a high-capacity Transformer \cite{vaswani2017attention} conditioned on the CVAE encodings, effectively capturing the variations and dependencies in the augmented dataset. Our policy also leverages the fact that robot movements are temporally correlated. Predicting action chunks \cite{act} instead of per-step actions, leads to smoother behaviors and mitigation of covariance shift commonly observed in the low data imitation learning regime. Next, we outline the individual components of our method in detail.

\subsection{Dataset (\roboset)}
\label{sec:roboset}
\begin{table}[t]
\centering
\caption{Open-source real-world manipulation dataset landscape: \roboset{}(ours) \url{https://robopen.github.io/roboset/} is one of the largest open-source robotics datasets. It contains high-quality demonstration, including human tele-operation, trajectories spanning a balanced distribution of 12 skills across 38 tasks in diverse kitchen scenes.}
\label{tb:datasets}
\resizebox{.8\columnwidth}{!}{%
\begin{tabular}{@{}llllll@{}}
\toprule
 & \textbf{Trajectories} & \textbf{Tasks} & \textbf{Skills} & \textbf{Scenes}  &\textbf{Source} \\ \midrule

\textbf{{\roboset} (\algoName)}   & \ntraj    & \ntask    &  \nskill  & 10    & TeleOp \\
\textbf{{\roboset} (kitchen)}    & 30,050    & 38    &  12   & 10      &  TeleOp \\ 
\textbf{{\roboset} (bin)}    & 70,000    & 10    &  4   & 1      &  Heuristics \\ 
\textbf{{\roboset} (full)}    & 98,050    & 48    &  12   & 11      &  TeleOp+Heuristics \\ 
\midrule

\ {BridgeData} \cite{ebertbridge}         & 33,200                 &        72        &  8                & 10        &  TeleOp \\
\ {BC-Z}~\cite{bcz}    &     25,000                  &          100      & 9               &             N/A   &  TeleOp \\
\ {RoboTurk} \cite{mandlekar2018roboturk}           & 2,100                 &       N/A         &   3     & 1     &      TeleOp \\ \midrule
\ {Amazon Pick-Place} \cite{mitash2023armbench} & 100,000               &      N/A          &       1          & 1       & Heuristics                 \\
\ {RoboNet} \cite{dasari2020robonet}             & 162,000               & N/A            &         2        & 7    &          Heuristics       \\
\ {BAIR Pushing}~\cite{bairpushing}    &     N/A                  &          N/A      & 1               &             1  &   Heuristics                          \\ \bottomrule
\end{tabular}
}
\end{table}

Training a general agent capable of robustly exhibiting a diverse repertoire of skills in novel scenes and tasks needs exposure to experiences matching this diversity. To align with our goal of building a data-efficient robot learning paradigm, we restrict ourselves to a frozen pre-collected small but diverse dataset -- \roboset{}(\algoName).
In order to capture behavioral diversity, we ensure sufficient coverage over different core skills, where each skill if defined as a temporally correlated sequence of actions that lead to plausible change in an object's pose. Example skills include \textit{closing/opening} articulated objects, \textit{sliding}, \textit{wiping}.
Each skill is instantiated across a set of objects. We refer to such (skill, object) combinations as a \textbf{task}. Our tasks are instantiated in different kitchen scenes, visually illustrated in \autoref{sec-app:dataset}. Instead of a random collection of tasks, we structure groups of tasks as belonging to be part of a household \textbf{activity}, such that they can be executed in sequence to obtain a meaningful outcome, such as cleaning a kitchen. 
Further details on the different skills, tasks and activities in \roboset{} are provided in Appendix~\ref{app-subsect:dataset-terminology}.

\label{sec:dataset}
\begin{figure}[h!]
    \centering
    \vspace{0em}
\includegraphics[width=0.9\textwidth]{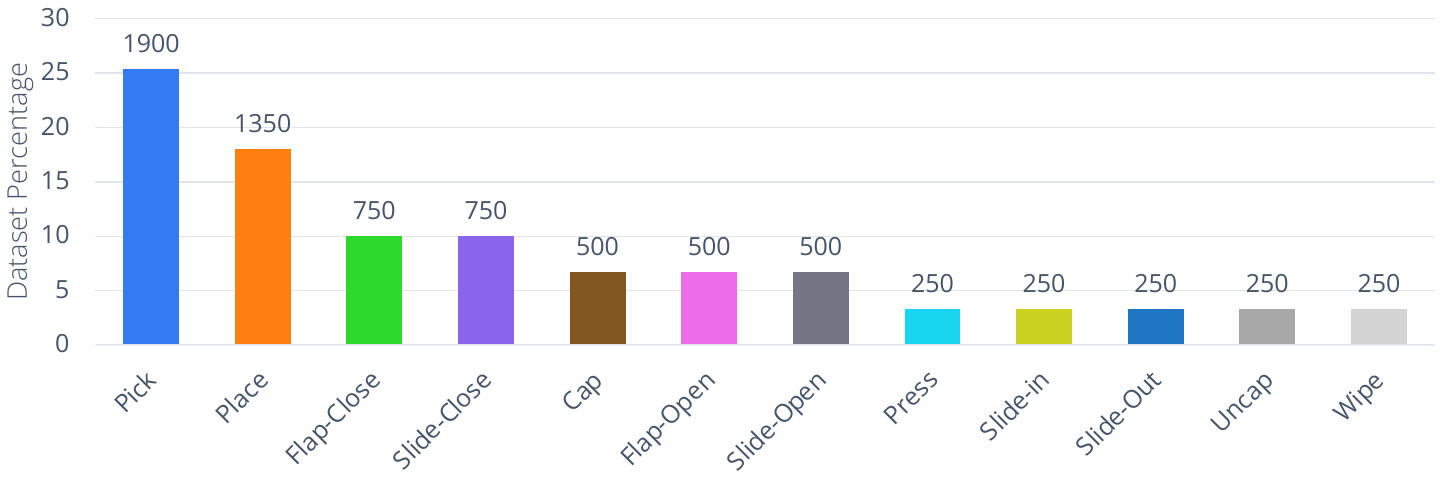}
    \vspace{0em}
    \caption{\footnotesize{\roboset{}(\algoName) contains a diverse set of 12 non-trivial manipulation skills (beyond picking/pushing, including articulated object manipulation and object re-orientation) expressed across 38 tasks across multiple scenes. In the figure, we mark the skill distribution of our dataset in terms of \% of trajectories with a certain skill. The number on top represents the number of trajectories corresponding to the skill used to train RoboAgent.}}
    \label{fig:skills}
\end{figure}

\texttt{RoboSet{}(\algoName{})} -- the dataset we used for this project consists of 7,500 trajectories (\autoref{tb:datasets})\footnote{Note that the entire \roboset{} is much larger and much more diverse. \agentName ~ is trained on \roboset(\algoName) -- a subset consisting of 7500 trajectories} collected using human teleoperation. The dataset involves \nskill skills expressed across multiple tasks and scenes. \autoref{fig:skills} shows the distribution of skills over our dataset.
While the commonly used pick-place skills cover 40\% of the dataset, we also include contact-rich skills such as (\emph{Wipe}, \emph{Cap}) as well as skills involving articulated objects (\emph{Flap-Open}, \emph{Flap-Close}).
We collect the overall dataset across four different physical setups. Each setup is instantiated with various everyday objects to create a kitchen scene. We frequently vary each set up with different variations of objects, thereby exposing each skill to multiple target objects and scene instantiations. We provide a glimpse of the overall setup and a subset of objects used in \autoref{fig:setup}, and data composition details in \autoref{sec:Data_Details}.

In \autoref{tb:datasets}, we compare our dataset with existing \emph{open-source} robot manipulation datasets.
Compared to prior open-source real-world datasets, \texttt{RoboSet} presents a greater number and diversity of skills and scene variations. \texttt{RoboSet} is one of the largest publically released dataset with commodity robots collected in the real-world setup.
Finally, despite our data diversity, \texttt{RoboSet(\algoName)} used for training our agents is still much smaller in size in comparison to other recent papers that use \emph{proprietary} robotic hardware, e.g. RT1 which has 135K trajectories \cite{rt1}, or those that rely on simulation~\cite{robocat}. 

\subsection{Data Augmentation}
\begin{figure}
    \centering
\includegraphics[width=0.99\textwidth]{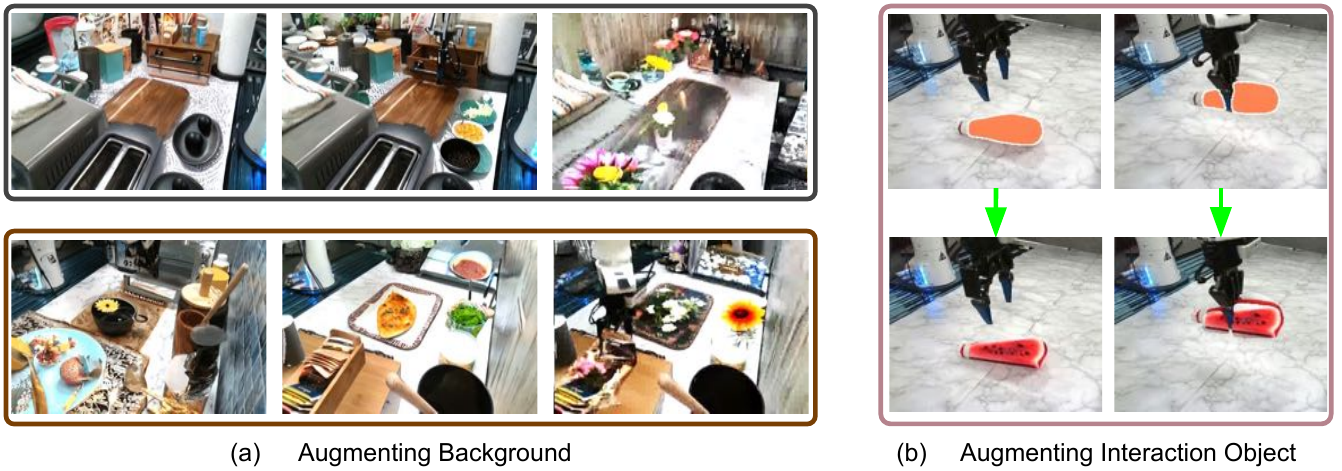}
    \caption{\footnotesize Illustration of the data augmentations that we develop to rapidly multiply limited robot datasets with diverse semantic scene variations. In (a) we show the scene around the robot and the interaction object changing. In (b), we show the interaction object itself changing while preserving the rest of the scene.} 
    \label{fig:aug-app}
\end{figure}

Generally useful robot manipulation systems will need to be able to deal with out-of-distribution scenarios (e.g. different homes and offices). Since any dataset of a practical size will have a limited diversity of objects and scenes (due to physical access and operational constraints) compared to what agents will encounter during deployment. To enable test-time generalization to the unseen scenarios, we develop a \emph{fully automatic} offline process to multiply the dataset.

Given an initial dataset of robot behaviors, we multiply the dataset by creating multiple semantic variations of the dataset while preserving the \textit{robot behavior} within each trajectory. These semantic variations are created by applying augmentations per frame within the trajectory. Augmentations are created by inpainting a part of the image frame introducing new objects and scene variations. The inpainting locations are specified by a mask and are informed by a text prompt. As opposed to \cite{cacti, genaug, rosie} needing manual masks, object templates, etc., our approach is fully automatic. We use the SegmentAnything model~\cite{sam} to automatically detect semantic boundaries in the scene to create augmentation masks. We apply augmentations separately to the object under manipulation and the rest of the environment respecting the object and robot boundaries. See \autoref{sec:semantic_aug} for additional details.
We emphasize that our approach toward semantic augmentation is fully automatic and offline. It takes advantage of and is also well poised to continually benefit from rapidly advancing progress in segmentation and in-painting models~\cite{sam,trackanything}. Akin to fields of natural language processing and computer vision, by distilling semantic real-world priors present in internet images/videos into robotics datasets, it provides robot learning a scalable mechanism to benefit from internet-scale data at no extra cost to humans/robots.

\vspace*{-0.5em}
\subsection{\algoName\space Architecture}
\label{sec:architecture}
\vspace*{-0.5em}

Scaling up dataset diversity as well as network capacity constitutes the two fundamental requirements to improve generalization in machine learning paradigms. In the field of robot learning, the need for generalization remains at large as both of the aforementioned generation requirements are hard to meet under real-world constraints. Safety and operational costs challenge the availability of diverse datasets, and the need for fast real-time control loops restricts the inference time needed to avail large models. 
Recovery of a generalizable robot manipulation policy under a practical data budget available in robotics demands an efficient policy architecture. In scenarios that have sufficient coverage within the training data, we want the policy to stay close to nominal behaviors (efficient imitation). The policy also needs to be effective to new variations (effective generalization) and contexts (efficient task conditioning) that are unseen during training . In addition, we want the policies to exhibit temporally correlated smooth behaviors accomplishing tasks with minimal errors and safety violations.

\begin{figure}[h]
    \centering
\includegraphics[width=0.95\textwidth]{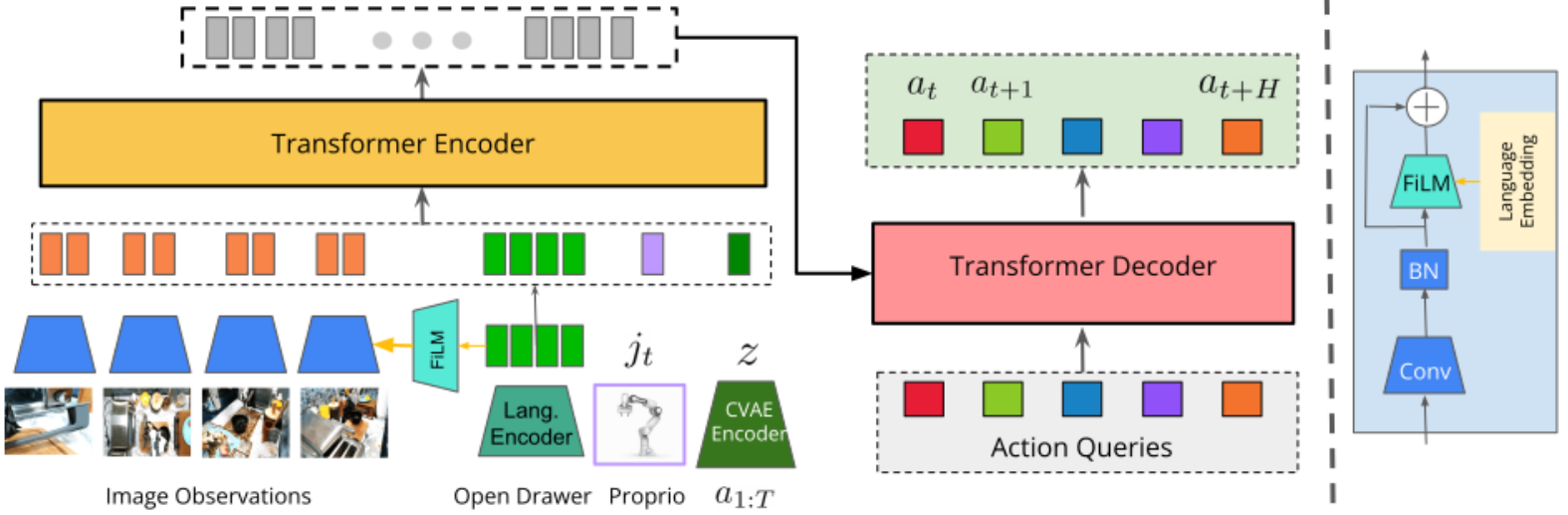}
    \caption{\footnotesize Policy architecture for \algoName{} . We use a CVAE that learns latent encodings $z$ for action sequences to implicitly identify different \textit{modes} in the data. A transformer takes as input a latent code, language embedding of the task, and image embeddings from four camera views, to autoregressively output an action sequence $a_{t:t+H}$ for chunk size $H$. On the right, we shows details for the FiLM layer \cite{perez2018film} that we use for language-conditioning.}
    \label{fig:architecture}
\end{figure} 

Our policy architecture -- \algoName{} is designed to be a Transformer model~\cite{vaswani2017attention} of sufficient capacity that can handle multi-modal multi-task robot datasets. In order to capture multi-modal data, following prior works~\cite{act} we incorporate a CVAE~\cite{kingma2013vae} that encodes action sequences into latent \textit{style} embeddings $z$. The decoder of the CVAE is the Transformer policy that conditions on the latents $z$. This formulation of expressing the policy as a generative model helps in effectively fitting to the multi-modal teleop data, without ignoring regions of a trajectory crucial for precision, which are also likely to be more stochastic. In order to model multi-task data, we incorporate a pre-trained language encoder~\cite{gadre2022clip} that learns an embedding $\mathcal{T}$ of a particular task description. To mitigate issues of compounding error and to achieve smooth temporally correlated robot motions, at each time-step, we predict actions $H$ steps in the future and execute them through temporal-aggregation of overlapping actions predicted for a particular time-step~\cite{act}. To improve effectiveness towards scene variations and robustness towards occlusions in clutter, we provide the policy with four different views of the workspace through four cameras. 

At time-step $t$, the transformer encoder takes four camera views , $o^{1:4}_t$, the joint pose of the robot $j_t$, the style embedding from the CVAE $z$, and the language embedding $\mathcal{T}$. We use a FiLM-based conditioning~\cite{perez2018film,rt1}, in order to ensure that the image tokens are able to reliably focus on the language instruction, such that the policy doesn't get confused about the task when multiple tasks are possible in a scene. The encoded tokens go to the decoder of the Transformer policy with fixed position embeddings, which finally outputs the next action chunk ($H$ actions) for the current time-step. For execution, we average over all overlapping actions predicted for the current time-step (As $H>1$, the action chunks overlap), and execute the resulting averaged action.     

\section{Experimental Design}
\label{sec:experiments}
Through experiments, we want to understand the following research questions

\begin{itemize}[leftmargin=*]
\itemsep0em
    \item How does \algoName{} perform, quantitatively and qualitatively, on a large set of vision-based robotic manipulation tasks? How does it generalize to new tasks, objects, and environments? 
    \item Does data augmentation improve robustness to noise/distractors?
    \item Does data augmentation improve policy generalization (i.e. scenes with new target objects)?
    \item Does the policy architecture of \algoName{} enable efficient learning with high performance?
        \item Does action chunking help with temporally consistent trajectories, achieving higher success?
\end{itemize}

To answer these research questions we instantiate our framework in the real world using commodity hardware and objects commonly used in everyday kitchens.  Next, we outline the system and dataset used to investigate our questions and then describe the different generalization axes for evaluation.

\begin{figure}[h!]
    \centering
\includegraphics[width=1.0\textwidth]{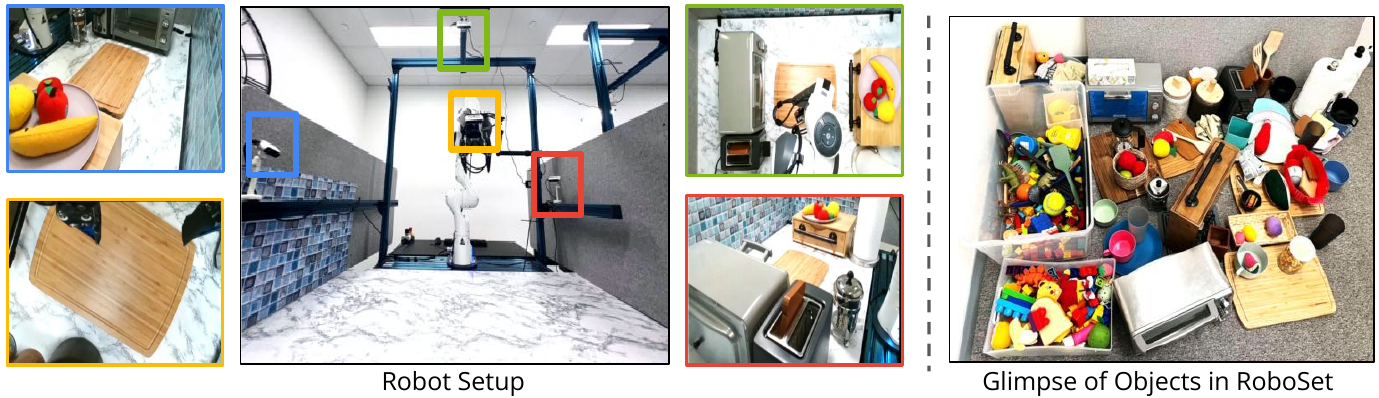}
    \vspace{0em}
    \caption{\footnotesize A zoomed-out view of the robot environment, showing all four cameras in the scene (in colored squares) that provide complementary views of the scene and a glimpse of the diverse set of objects in the dataset, used for all our experiments. The objects range from articulated objects like drawers and ovens, to smaller rigid objects like french press, bowls, and deformable objects like towels.}
    \label{fig:setup}
\end{figure}

\textbf{Robot hardware.}  \autoref{fig:setup} shows our robot environment, called \textit{RoboPen} that consists of a kitchen setup with everyday objects, a Franka Emika Panda arm with a two-finger Robotiq gripper fitted with Festo Adaptive Fingers\footnote{\url{https://robotiq.com/products/2f85-140-adaptive-robot-gripper} and \url{https://www.festo.com/us/en/p/adaptive-gripper-finger-id_DHAS_GF/}}, three fixed  cameras (top, left, right), and a wrist camera mounted above the end-effector.  The four Realsense D455 camera views provide complementary perspectives of the workspace, and we utilize all of them for robust policy learning.

\label{sec:Data_Details}
\textbf{Data collection.} Our robot manipulation dataset for the experiments consists of \ntraj trajectories, collected through tele-operation by a human operator, over a period of two months. We collect all the data across four different physical setups with different kitchen-like environments with a Franka Emika~\cite{haddadin2022franka}. Each setup also sporadically changes its scene by swapping new objects and backgrounds. The teleoperation stack is based on~\cite{kumar2015mujoco} and uses VR-controllers. The dataset comprises of diverse manipulation skills like opening/closing drawers, pouring, pushing, dragging, picking, placing, etc. across several everyday objects. \autoref{fig:skills} shows the distribution of skills in the dataset. Additional details regarding the dataset, along with sample trajectories, and a link to the entire dataset are in the \href{https://robopen.github.io/}{project website\extweblink} linked with the paper.
We are publicly releasing this dataset, as part of RoboSet -- a large multi-skill robotics dataset described in \autoref{sec:roboset}. To our knowledge, this is one of the largest open-source robot manipulation datasets with the most commonly used non-proprietary robot hardware (Franka Panda~\cite{haddadin2022franka}) containing diverse real-world behaviors beyond pick and place. 

\begin{figure}[h!]
    \centering
     \includegraphics[width=0.95\textwidth]{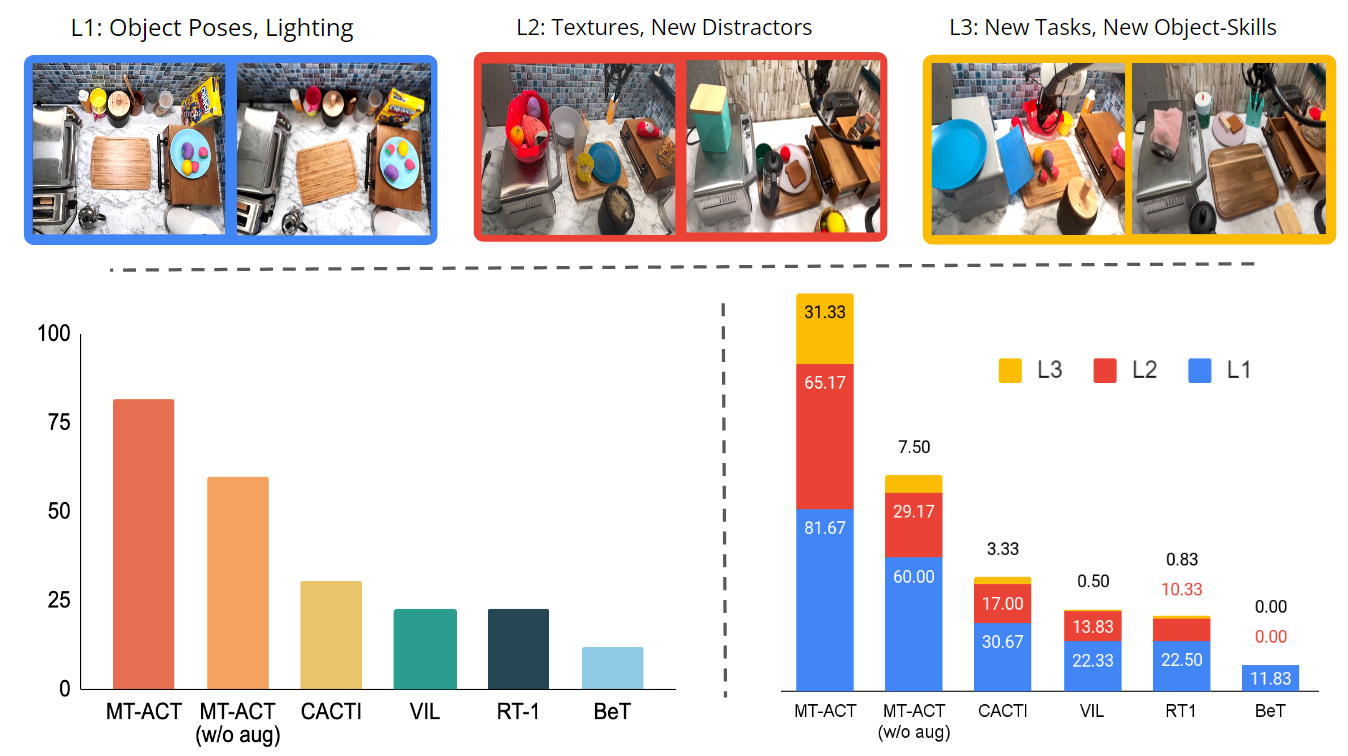}
    \caption{\footnotesize Visualization of different generalization axes, evaluating effectiveness with lighting variations and smaller scene changes such as object poses (L1), robustness to significant scene variations (L2), generalization to unseen tasks (L3). \emph{Bottom-Left:} Results for commonly used L1-generalization. \emph{Bottom-Right:} Multi-Task (universal policy) results for different levels of generalization. See~\ref{fig:l4_generalization} for L4-generalization results.}
    \label{fig:gen_vis}
\end{figure}
\textbf{Generalization Axes.}
\label{sec:gen_axes} Following prior work~\cite{rt1,bcz,vima}, we define each \textit{task} to consist of a particular language command like \textit{`pick a cube of butter from the drawer on the left'} that defines an object to be interacted with (butter), a skill to be executed (pick), and some context (drawer on the left). Each activity consists of a collection of 4-5 closely tasks that can be executed in sequence. Table~\ref{tab:activity-table} lists the different activities used in our work. We refer to the policy trained over all the activities to be the \textit{universal policy}, and those trained only over one activity to be an \textit{activity policy}. We consider evaluations in terms of different levels of generalization, illustrated visually for a scene in Fig.~\ref{fig:gen_vis}:  \textbf{L1(Effectiveness)}: Generalization of the agent to variations in object positions and orientations, and in lighting conditions. \textbf{L2 (Robustness):} New background, different distractor object variations, and unseen distractor objects introduced in the scene. \textbf{L3 (Generalization):} New tasks never seen before, including new object-skill combinations. \textbf{L4 (Strong Generalization):} New kitchen never seen before (see Figure~\ref{fig:l4_generalization} Left).

\section{Experiments}

Through detailed real-world robot manipulation experiments, we evaluate the proposed framework for sample efficiency, and generalization of the agent to diverse scenes. 

\textbf{Baselines.}
We compare multiple baselines that use visual policy learning for robotics.  \textit{Single Task Agents}: We compare ACT-based policies~\cite{act} trained for individual tasks, and evaluated on the respective tasks. These policies don't need to generalize across tasks and scene, and represent an approximate \textit{oracle} performance per task.\textit{ Visual Imitation Learning (VIL)}: We compare with regular language-conditioned multi-task visual imitation learning. \textit{CACTI}~\cite{cacti}: This baseline is a prior framework for multi-task learning that also uses some scene augmentations for generalization. \textit{RT1}~\cite{rt1}: We re-implement a baseline RT1-like agent. \textit{BeT}~\cite{bet}: We modify the Behavior Transformer architecture with language conditioning and train it in a multi-task manner.

Next, we present results and analysis from our large-scale real-world experiments that attempt to understand the research questions presented in section~\ref{sec:experiments}. 

\subsection{Multi-Task Real-World Results}

\textbf{Performance.}
Figure~\ref{fig:gen_vis} (Left-Bottom) compares our proposed MT-ACT policies against commonly used imitation learning architectures.
In this figure (Figure~\ref{fig:gen_vis} Left-Bottom) we only plot results for \emph{L1-generalization} since this is the standard setting most other imitation learning algorithms use.
We observe that all approaches that only model next-step actions (instead of sub-trajectories) exhibit weaker performance.
Among these approaches, we find that action-clustering-based approaches (BeT~\cite{bet}) for multi-task settings, perform significantly worse. We believe this happens because naive clustering in very diverse action distributions may not result in clusters that generalize across diverse skills.
Additionally, since we are operating under a data budget, we observe that RT1-like approaches that require a lot of data do not perform well in the low data regime.
By contrast, our \algoName{} policy which uses action-chunking and CVAE to model multi-modal sub-trajectories  significantly outperforms all baselines.

\begin{figure}[t]
    \centering
    \includegraphics[width=\textwidth]{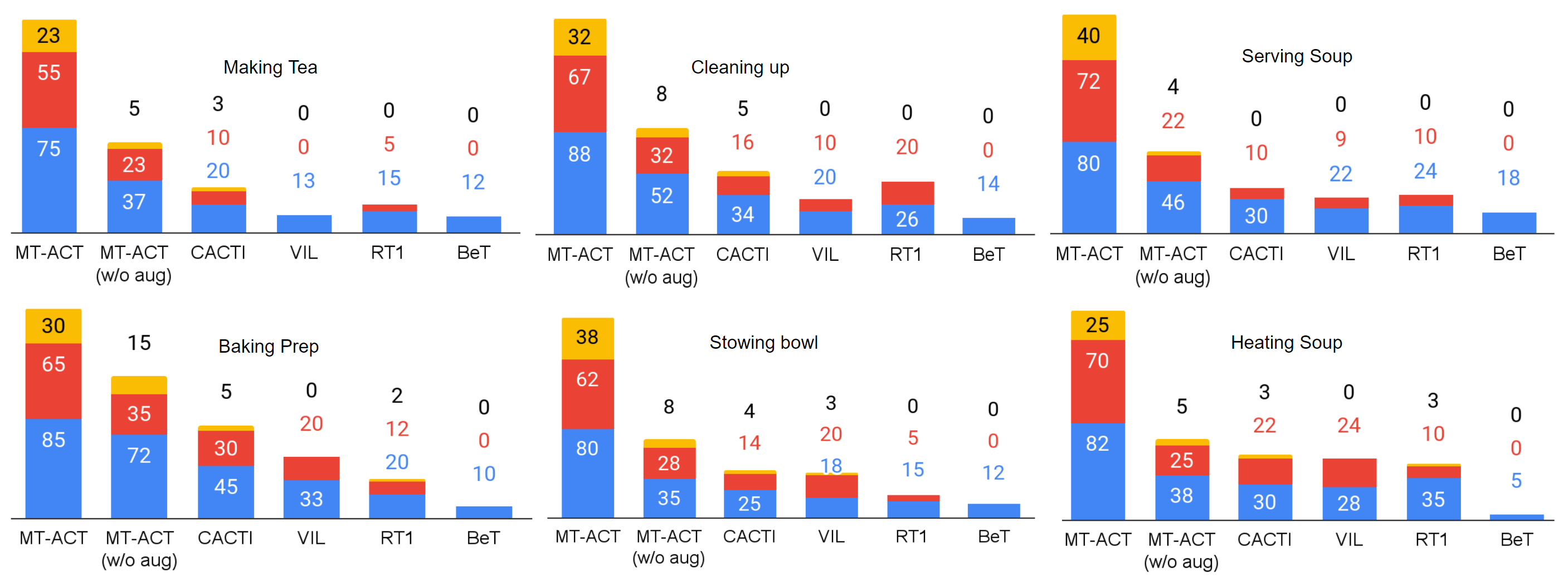}
    \caption{\footnotesize Results for \algoName, its ablated variant without semantic augmentations, and baselines, for different activities, with L1, L2, L3 levels of generalization. Each activity consists of 4-5 tasks, and the results average over the tasks in an activity. The results show that semantic augmentations significantly improve performance of MT-ACT  over the baselines. In addition, even without augmentations, the MT-ACT policy achieves much higher success rates compared to the baselines.}
    \label{fig:mainresults}
\end{figure}

\textbf{Generalization and Robustness.}
Figure~\ref{fig:gen_vis} (Bottom-Right) shows the results for all methods across multiple levels of generalization (\textbf{L1}, \textbf{L2}, and \textbf{L3}).
Recall that these levels of generalization include diverse table backgrounds, distractors (\textbf{L2}) and novel skill-object combinations (\textbf{L3}).
From Figure~\ref{fig:gen_vis} (Bottom-Right) we see that by virtue of semantic augmentations and action representations, \algoName{} significantly outperforms all the baselines we consider. 
More interestingly, we see that semantic augmentations have less effect for L1-generalization ($\approx 30\%$ relative), they provide a \emph{much more} significant improvement for both L2-generalization ($\approx 100\%$ relative) and L3-generalization ($\approx 400\%$ relative).
Since semantic augmentations affect both  scenes (backgrounds and distractor objects) as well as target objects being manipulated they provide useful support for the policy to achieve increasing levels of generalization.

Additionally, in Figure~\ref{fig:mainresults} we also separately report generalization results for each activity separately. 
From Figure~\ref{fig:mainresults} we see that each our proposed semantic augmentations positively affect each activity's performance.  
Interestingly, we find that for some of the harder activities (Making-Tea, Stowing-Bowl, Heating Soup) the relative improvement in performance due to semantic augmentations is much larger. 
\begin{wrapfigure}{r}{0.40\textwidth}
    \centering
  \includegraphics[width=0.40\textwidth]{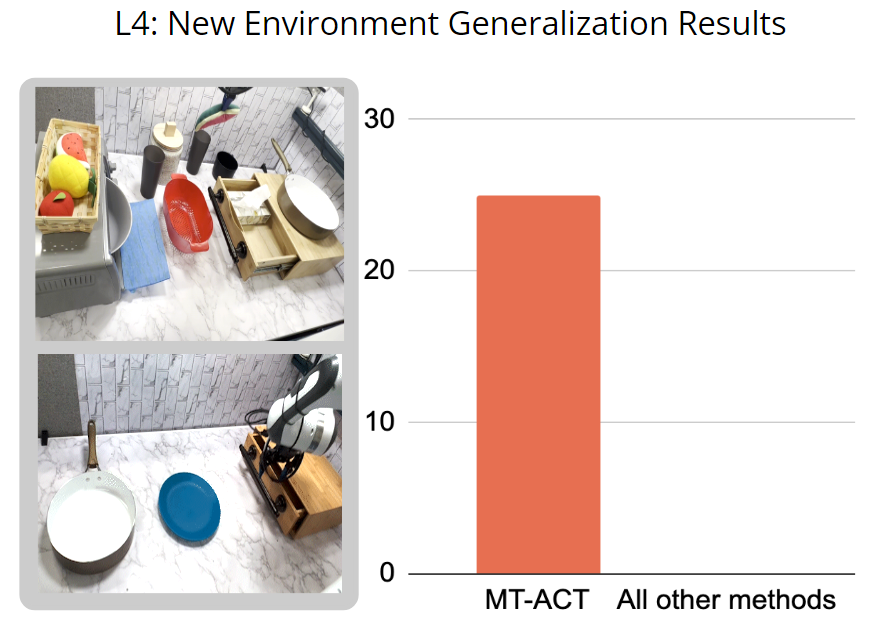}
    \caption{\footnotesize Only MT-ACT policies perform tasks in a completely new kitchen environment (L4).}
    \label{fig:l4_generalization}
\end{wrapfigure}
Overall, our results show that traditional visual imitation learning (without any augmentations), i.e., VIL and RT1 trained on our relatively small dataset, completely fail for L3 and L2,
indicating a lack of generalization to unseen scenarios, due to data paucity.  
Finally, we also test our policy on a completely new kitchen with novel objects, arrangements, distractors, i.e., L4 generalization.
Figure~\ref{fig:l4_generalization} (Left) visualizes this new kitchen environment.
We evaluate all methods in this new kitchen for 3 tasks. 
Figure~\ref{fig:l4_generalization} (Right) shows the results for each method on this new environment. Specifically, we obtain an average success rate of 25\% for \algoName{} (and 0 for all the other baselines). 
We see that \algoName{} without semantic augmentations also fails completely on this new environment thus indicating the benefits of our approach for zero-shot adaptation.

\subsection{Ablation}
\label{sec:ablation} We ablate the different design choices we make in our proposed architecture.

\begin{figure}[t]
    \centering
    \includegraphics[width=0.99\textwidth]{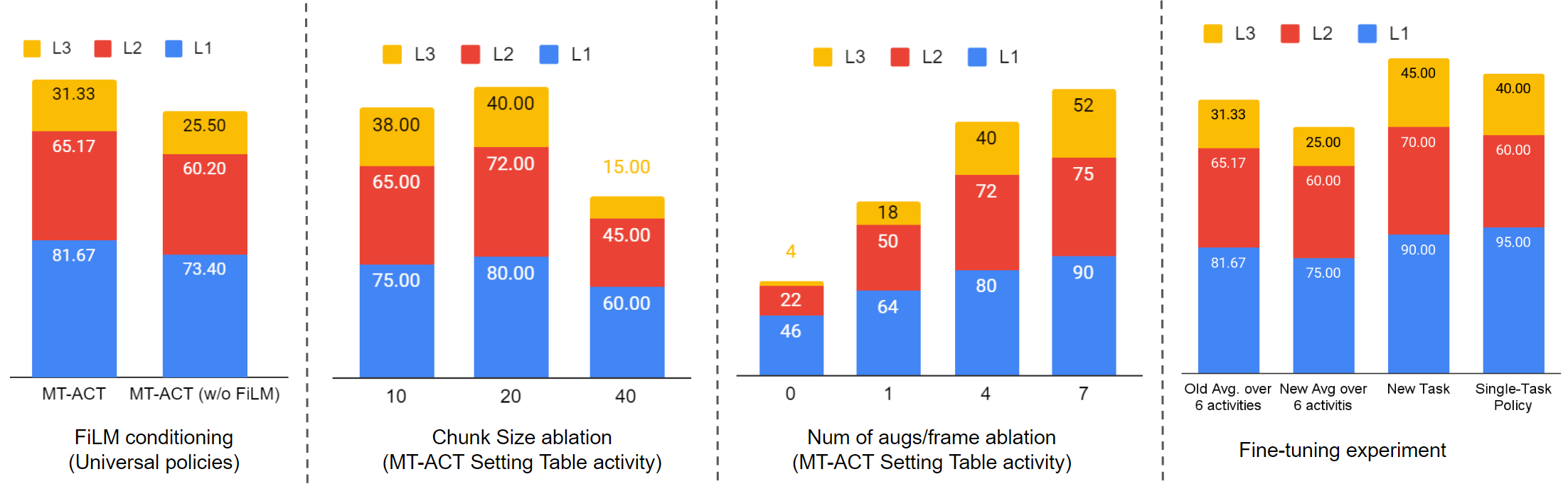}
    \caption{\footnotesize We shows results of the different ablation studies and analysis in section~\ref{sec:extension}, showing the benefits of FiLM conditioning, the effect of varying chunk sizes in the predictions, the number of augmentations per frame for multiplying the dataset, and the feasibility of fine-tuning \algoName{} for improved deployment.}
    \label{fig:ablation}
\end{figure}

\textbf{Task Specification using FiLM conditioning.} For language conditioned multi-task policy, as described in section~\ref{sec:architecture}, we use a FiLM based conditioning~\cite{perez2018film} for the language embedding of task descriptions~\cite{bert}. Here, we compare this design choice with a simple concatenation-based conditioning of the language embeddings with image tokens for the policy.  In Fig.~\ref{fig:ablation} (Left) we show results for this ablation study averaged over all activities, and we observe a 5-10\% drop in performance of the version of \algoName{} without FiLM conditioning.

\textbf{Chunk Size for Action Representations.} Here we train variants of \algoName{} with different chunk sizes 10, 20, 40. In Fig.~\ref{fig:ablation} ((Middle-Left), we see that a chunk size of 20 performs the best, with a 0-5\% drop in performance with chunk size 10. In addition, large chunk size 40 performs significantly worse with more than 20\% drop in performance indicating the inability of the policy to correct errors as the chunks grow in size.

\textbf{Number of augmentations per frame.} We ablate the number of augmentations per frame, to see if more augmentations help \algoName{} in learning a more performant policy. From Fig.~\ref{fig:ablation} (Middle-Right), we see that the number of augmentations per frame is strongly correlated with overall performance gains. Thanks to the real-world semantic priors injected via data augmentation, the gains are more notable for L2 and L3 levels where out-of-domain generalization is required from the policy.

\textbf{Robustness analysis.} We perform several robustness analyses of the universal \algoName{} agent, by manually perturbing the scene during evaluation, and also introduce system failures like blocking the views from one, two, or three cameras. On average, we find that the policy is robust to these strong active variations, and can solve the specified task in about 70\% of the 20 evaluations we run for this analysis. Videos showing these results are in the project website.

\textbf{Plasticity.}
\label{sec:extension}
Here, we evaluate the feasibility of adding additional capabilities to the universal \algoName{} agent, without requiring significant re-training. We take the trained agent (on \ntask tasks) and fine-tune on $(1/10)^{\text{th}}$ of the original data combined with data for a new held-out task (placing toast in toaster oven). The new task has 50 trajectories, multiplied with 4 augmentations per frame, for a total of 250 trajectories. Fig.~\ref{fig:ablation} (Right) shows that the fine-tuned agent is able to learn this new task, without significantly deteriorating in performance on the previous \nactivity activities. Also, it achieves slightly better L2, L3 performance than a single-task policy trained only on augmented data of the new task, indicating efficient data re-use.

\section{Discussion and Limitations} 
We developed a framework for sample-efficient and generalizable multi-task robot manipulation in the real world. Our framework is based on rapidly multiplying a small robotics dataset through semantic scene augmentations, and training a multi-task language-conditioned policy that can ingest the diverse multi-modal data obtained through augmentations. We combine and adapt several design choices like action chunking and temporal aggregation proposed in the context of single-task policies, and show that they yield significant boosts in performance even in the multi-task settings we consider. 

Finally, we release one of the largest robot manipulation datasets to date involving over \nskill skills in kitchen environments which we hope will facilitate further research in developing robot manipulation systems with diverse real-world generalization. An important limitation of our work is that all the tasks consist of individual skills, and an interesting direction for future work would be to develop approaches for composing skills automatically for solving long-horizon tasks. Another limitation is that we do not explore the axes of language generalization, and use language embeddings from pre-trained encoders as is, without any modifications. Future work could investigate better language conditioning that is more flexibly adaptable to changes in task descriptions.

\section*{Acknowledgements}
We acknowledge various contributions, large and small, from the entire Embodied AI team at Meta.
The project has also significantly benefitted from brainstorming sessions from -- Aravind Rajeswaran, Chris Paxton, Tony Zhao, Abhishek Gupta, and individual contributions from Giri Anantharaman, Leonid Shamis, Tingfan Wu, Priyam Parashar, Chandler Meadows, Sahir Gomez, and Liyiming Ke. We thank Gaoyue Zhou, Raunaq Bhirangi, Sudeep Dasari , Yufei Ye, Mustafa Mukadam, Shikhar Bahl, Mandi Zhao, Wenxuan Zhou, Jason Ma, and Unnat Jain for helpful discussions at different stages of the project.

\bibliography{example}  %

\newpage
\clearpage
\appendix
\section{Dataset details}
\label{sec-app:dataset}

\algoName{} uses 7,500 human teleoperated demonstrations from the \emph{RoboSet} dataset \footnote{The full RoboSet is much more diverse and consists of 9,500 teleoperated demonstrations, 20,500 kinesthetic demonstrations in various kitchen and table-top settings. In addition, it contains about 70,000 trajectories in bin settings collected through heuristics }. \algoName{} dataset consisted of RGB and depth frames from four camera views (right, left, top, and wrist) as shown in figure \ref{fig:setup}, Franka joint positions and velocities, end-effector/gripper position and velocities, controls applied to the Franka joints and end-effector/gripper, and the time-steps (40 steps). 

\begin{figure}[h!]
    \centering
    \vspace{0em}
\includegraphics[width=1\textwidth]{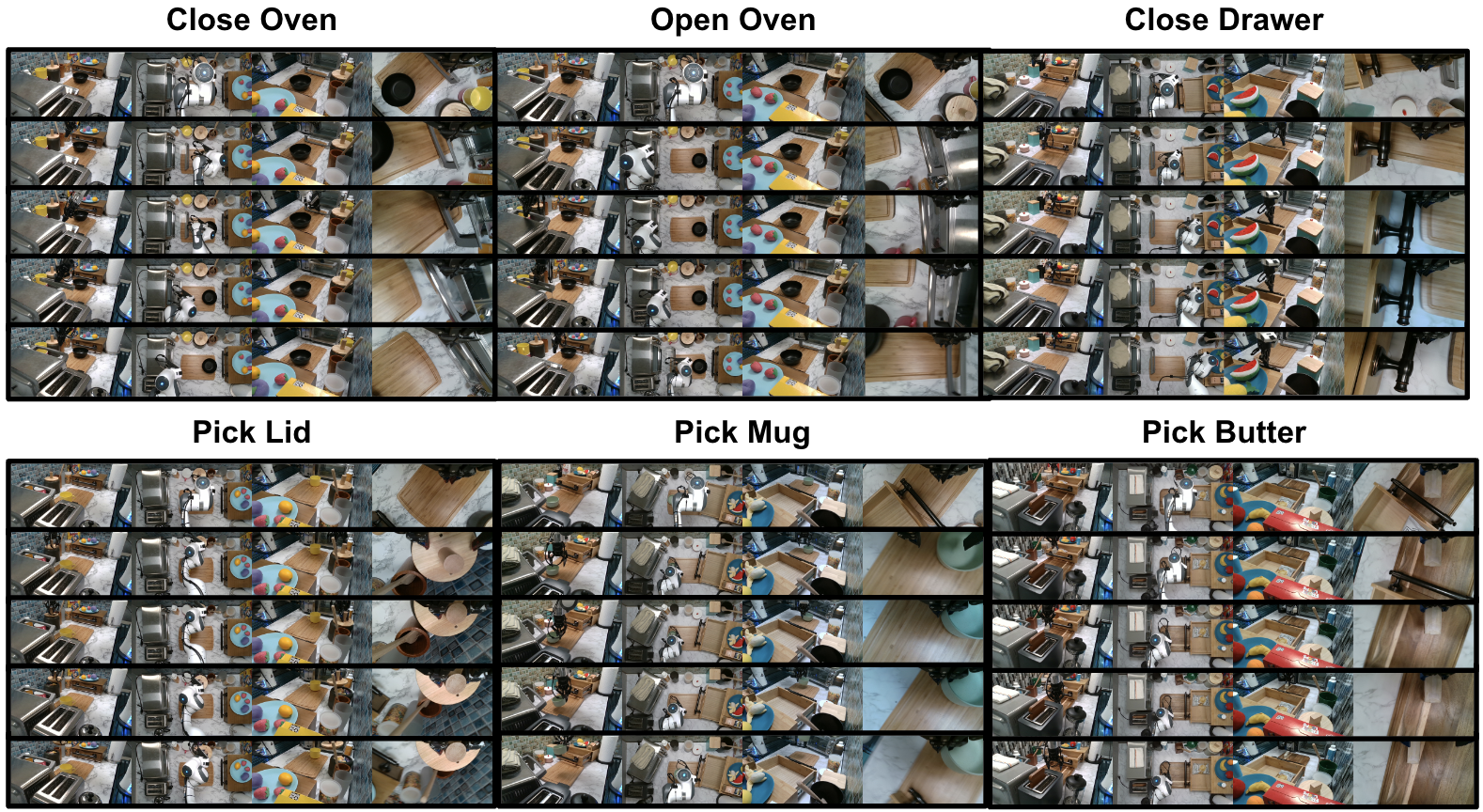}
    \vspace{-1.5em}
    \caption{\footnotesize Sample task demonstrations in the \roboset (visualizing four views horizontally, and five timesteps vertically), used for training.}
    \label{fig:rollouts}
\end{figure}

The data was collected using an Oculus Quest 2 controller on a kitchen table-top setup at 5Hz and saved in HDF5 format. Rollouts from the data are shown in Figure~\ref{fig:rollouts} as well as in \url{https://robopen.github.io/roboset/}.

\begin{table}[]
\resizebox{\textwidth}{!}{%
\begin{tabular}{@{}llllllllll@{}}
\toprule
\rowcolor[HTML]{fefae0} 
Heat Soup &  & Serve Soup &  & Baking Prep & Making Tea &  & Cleaning Up &  & Stow Bowl \\ \midrule
Flap-Open Oven &  & Flap-Open Oven &  & Slide-Open Drawer & Uncap Lid &  & Pick Lid &  & Slide-Open Drawer \\
Pick Bowl &  & Pick Bowl &  & Pick Butter & Place Lid &  & Cap Lid &  & Pick Bowl \\
Slide-In Bowl &  & Slide-Out Bowl &  & place Butter & Pick Tea &  & Slide-Close Drawer &  & Place Bowl \\
Flap-Close Oven &  & Flap-Close Oven &  & Slide-Close Drawer & Place Tea &  & Flap-Close Oven &  & Slide-Close Drawer \\
 &  &  &  &  & Pick Lid &  & Pick Towel &  &  \\ \bottomrule \\
\end{tabular}%
}
\caption{List of activities (Top Row) and the associated tasks for each activity.}
\label{tab:activity-table}
\end{table}

\subsection{Dataset Terminology}
\label{app-subsect:dataset-terminology}

\textbf{Skill}
Different works in robotics often assign different meanings when they refer to ``skills''.
In our work, we refer to a skill when the robot performs a similar motion across different object instances (both shape and size). For instance, pick, place, open, close objects are considered as different skills. Since our dataset contains articulated objects if the ``open'' skill with multiple objects results in different motion we classify them as different skills.
For instance, ``Open Drawer'' requires interacting with a prismatic joint while ``Open Oven'' interacts with a revolute joint. Hence, we classify these as separate skills. Our definition is broadly similar to some previous works \cite{rt1}. 
We use 12 skills in RoboSet --
\emph{Slide-Open, Slide-Close, Flap-Open, Flap-Close, Cap, Uncap, Pick, Place, Wipe, Plunge, Slide-in, Slide-out}.

\textbf{Task:} We define each instantiation of our skill with a particular object class as a different task. For instance, ``Pick Mug'' and ``Pick Butter'' correspond to the same ``Pick'' skill but are two different tasks.

\textbf{Activity:}
A general robot agent will eventually need to perform a sequence of tasks, e.g. make tea.
We refer to such sequence of tasks as \emph{activities}. 
Table~\ref{tab:activity-table} lists the activities used in our work as well as tasks corresponding to each activity.
Our final aim is to train a \emph{single} robot agent to perform all activities.

\textbf{Policies:}
We train and compare different policies in our work. We classify these policies into \emph{single-task} policy, \emph{multi-task (single-activity)} and \emph{multi-task (universal)} policies.
As each name suggests, single-task policies are trained on specific tasks.
Multi-Task (single-activity) policies are trained on all tasks belonging to an activity. 
Finally, Multi-Task (universal) policies are trained on all tasks and activities.
Our final \agentName{} is trained as a Multi-Task (universal) policy.

\subsection{Details on Semantic Augmentations}
\label{sec:semantic_aug}
We enable two different types of scene augmentations for multiplying data, for enabling generalization to different scenes with novel distractors, and to scenes with different objects for interaction: 
\begin{itemize}[leftmargin=*]
    \item \textbf{Augmenting interaction object:} Given the joint angle of the robot in a frame of a trajectory, we use forward kinematics to recover the robot mask as well as the end-effector position of the robot. We use the end-effector location to prompt SegmentAnything~\cite{sam} for obtaining a mask of the object being interacted with. We then inpaint the region of the object being interacted with, based on a text prompt, and keep it consistent across time by tracking with TrackAnything~\cite{trackanything}. 
        \item \textbf{Augmenting background:} We use SegmentAnything~\cite{sam} to randomly choose a set of objects in the background that do not overlap with the robot mask, and the mask of the object being interacted with, and inpaint the scene based on the resulting overall mask over all the objects identified by SegmentAnything. 
\end{itemize}

Note that our augmentation approaches are all automatic and do not require any manual effort in specifying masks or object meshes etc. This is in contrast to prior works that require manual specification of a fixed mask per trajectory~\cite{cacti}, and those that require templates of object textures and meshes~\cite{genaug}. In addition, unlike~\cite{rosie}, we do not require training any further modules for identifying objects through open-vocabulary detection that relies on language grounding. 

\section{Train and Evaluation Details}
In this section we present training and evaluation details both for our methods and the baselines.

\subsection{Robot Environment and Evaluation Details}
The robot environments for evaluation consist of table-top kitchen setups with diverse real objects in the scene. There are 4 cameras providing complementary views of the workspace. The robot is a  Franka Emika Panda arm operated with joint position control, with an action space dimension of 8 (7 joint positions, 1 dimension for end-effector open/close). The robot arm has a two-finger gripper, and a wrist camera. The robot is operated at a frequency of 5Hz. 
\begin{figure}[t]
    \centering
    \includegraphics[width=\textwidth]{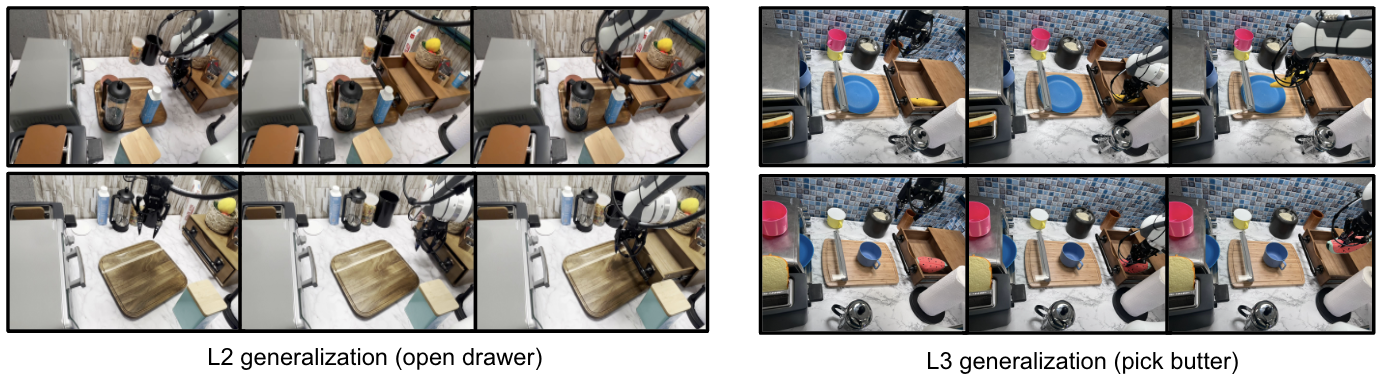}
    \caption{\footnotesize Qualitative results of rollouts for L2 and L3 levels of generalization, showing tasks \textit{open drawer} and \textit{pick a slab of butter}. For L2 we introduce different distractors in the scene, and change the background tiles. For L3, in addition to changes in L2 we introduce different task objects, for example by replacing a slab of butter with a piece of watermelon, or a banana.}
    \label{fig:qual}
\end{figure}

\subsection{Hyper-parameters for \algoName{} and baselines}
Here we provide hyper-parameter details of the policy architecture. We train all policies for 2000 epochs. For the overall \algoName{} agent, trained on the augmented dataset, this takes about 48 hours on a single 2080Ti GPU with a batch size of 8. 

For our baseline implementations we did a hyperparameter search for relevant parameters. 
For each baseline implementation we try to adapt them from their officially released code. Specificially, for RT1 \cite{rt1} we use \url{https://github.com/google-research/robotics_transformer} for reference. 
On the other hand, for BET \cite{bet} we use \url{https://github.com/notmahi/bet}.
To provide language conditioning for both baselines we use similar FiLM \cite{perez2018film} implementation as our approach.

For hyper-parameters we use 3 different discrete action sizes -- 64, 256 and 512,
we vary the learning rates from $(1e-3, 1e-4)$. We use the AdamW optimizer with a weight decay range in $(1e-2, 1e-3, 1e-4)$. Our RT-1 transformer uses 6 layers with 8 parallel attention heads and each head with size 64. Each transformer uses a feedforward layer with intermediate sie of 1024.
On the other hand for \cite{bet} we experiment with 3 different action cluster sizes -- 64, 256 and 512.
We use a similar transformer implementation for BET as RT-1.
Finally, for real-world evaluation we use the hyper-parameters with lowest validation loss. 

\begin{table}[t]
\begin{minipage}{.48\linewidth} 
\centering
\caption{Hyper-parameters for \algoName}
\begin{tabular}{@{}cc@{}}
\toprule
\textbf{Name}              & \textbf{Value} \\ \midrule
learning rate              & 1e-5           \\
batch size                 & 8              \\
feedforward size           & 3200           \\
Attention heads            & 8              \\
chunk size                 & 20             \\
dropout                    & 0.1            \\
Transformer encoder layers & 4              \\
Transformer decoder layers & 7              \\
Language Embedding size     & 384            \\ \bottomrule
\end{tabular}
\end{minipage}
\hfill
\begin{minipage}{.48\linewidth}
\centering
\caption{Hyper-parameters for RT-1 \cite{rt1}}
\begin{tabular}{@{}cc@{}}
\toprule
\textbf{Name}              & \textbf{Value} \\ \midrule
learning rate              & 1e-4           \\
discrete action tokens & 256           \\
batch size                 & 64         \\
feedforward size           & 1024           \\
Attention heads            & 8              \\
dropout                    & 0.1            \\
Transformer layers & 6              \\
Language Embedding size     & 384            \\ \bottomrule
\end{tabular}
\end{minipage}
\end{table}

\begin{figure}[t]
    \centering
    \includegraphics[width=0.94\textwidth]{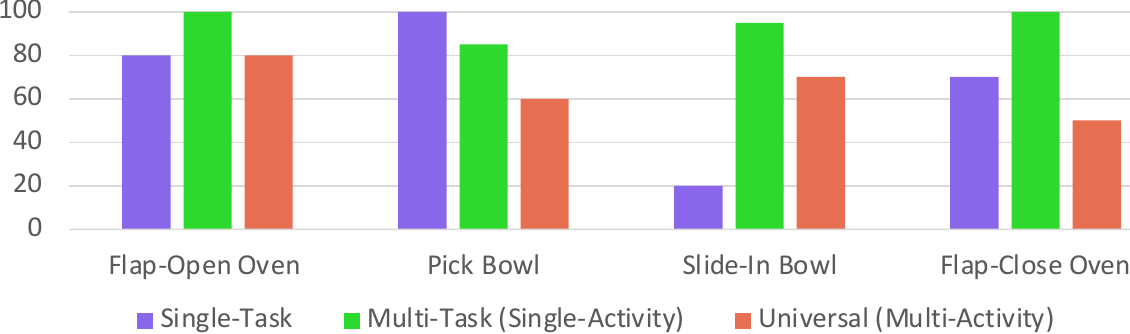}
    \caption{\footnotesize Single-Task vs Multi-Task comparison for Heat Soup activity. Multi-Task (Single Activity) represents a multi-task policy trained on only 4 tasks in Heat-Soup activity.}
    \label{fig:single-task-vs-multi-task}
\end{figure}

\section{Additional Results}

In this section, we present some additional results.
First, we present results and discuss how well our multi-task policy performs when compared to single-task policies.
Figure~\ref{fig:single-task-vs-multi-task} compares single-task policy performance against two sets of multi-task policies for the \emph{Heat Soup} activity.
For the first multi-task Single-Activity policy (MT Single-Activity) we only train it across all tasks within the same activity.
For the latter multi-task universal multi-activity policy (MT-Universal) we train it across all tasks in all activities. 
From Figure~\ref{fig:single-task-vs-multi-task} we see that for most tasks \emph{MT Single-Activity} is able to outperform single task policies.
Additionally, single-task policies are able to perform well on most tasks ($\approx 80\%$) except
the more challenging constrained manipulation tasks (slide-in-bowl) ($\approx 20\%$).
Finally, we also observe that MT-Single-Activity can outperform MT-Universal for most tasks. 
This happens because the universal agent is trained to perform a much larger variety of tasks. Given the very large variety of skills (Figure~\ref{fig:skills}),
such multi-task training can result in some negative transfer compared to training on a narrowly defined skills.
We believe these reduced multi-task results present useful avenues for future research.
Finally, in Table~\ref{tab:universal-all-tasks} we show results for all tasks in all activities using our single universal policy. 
From the below table, we see that the universal policy is able to perform well on most tasks except the more challenging tasks such as grasping small deformable objects (Pick Tea: 40\%, Pick Lid: 50\%).

\begin{table}[b]
\resizebox{\textwidth}{!}{%
\begin{tabular}{@{}llllllll@{}}
\toprule
\textbf{Heat Soup} & Success &  & \textbf{Serve Soup} & Success &  & \textbf{Baking Prep} & Success \\ \midrule
Flap-Open Oven & 80\% &  & Flap-Open Oven & 90\% &  & Slide-Open Drawer & 70\% \\
Pick Bowl & 60\% &  & Pick Bowl & 50\% &  & Pick Butter & 70\% \\
Slide-In Bowl & 70\% &  & Slide-Out Bowl & 80\% &  & place Butter & 90\% \\
Flap-Close Oven & 50\% &  & Flap-Close Oven & 80\% &  & Slide-Close Drawer & 90\% \\ 
& & & & & & & \\ \midrule
\textbf{Making Tea} & Success &  & \textbf{Cleaning Up} & Success &  & \textbf{Stow Bowl} & Success \\ \midrule
Uncap Lid & 80\% &  & Pick Lid & 70\% &  & Slide-Open Drawer & 70\% \\
Place Lid & 90\% &  & Cap Lid & 100\% &  & Pick Bowl & 70\% \\
Pick Tea & 40\% &  & Slide-Close Drawer & 90\% &  & Place Bowl & 80\% \\
Place Tea & 60\% &  & Flap-Close Oven & 80\% &  & Slide-Close Drawer & 80\% \\
Pick Lid & 50\% &  & Pick Towel & 90\% &  &  &  \\
Cap Lid & 70\% &  & Wipe Counter & 90\% &  &  &  \\ \bottomrule \\
\end{tabular}%
}
\caption{Results for different tasks using the learned \textbf{universal policy}.}
\label{tab:universal-all-tasks}
\end{table}

\end{document}